# LARGE LANGUAGE MODELS AND THE REVERSE TURING TEST

Terrence J. Sejnowski

terry@salk.edu
Salk Institute for Biological Studies
University of California, San Diego

Large Language Models (LLMs) have been transformative. They are pre-trained foundational models that are self-supervised and can be adapted with fine-tuning to a wide range of natural language tasks, each of which previously would have required a separate network model. This is one step closer to the extraordinary versatility of human language. GPT-3 and more recently LaMDA can carry on dialogs with humans on many topics after minimal priming with a few examples. However, there has been a wide range of reactions and debate on whether these LLMs understand what they are saying or exhibit signs of intelligence. This high variance is exhibited in three interviews with LLMs reaching wildly different conclusions. A new possibility was uncovered that could explain this divergence. What appears to be intelligence in LLMs may in fact be a mirror that reflects the intelligence of the interviewer, a remarkable twist that could be considered a Reverse Turing Test. If so, then by studying interviews we may be learning more about the intelligence and beliefs of the interviewer than the intelligence of the LLMs. As LLMs become more capable they may transform the way we interact with machines and how they interact with each other. Increasingly, LLMs are being coupled with sensorimotor devices. LLMs can talk the talk, but can they walk the walk? A road map for achieving artificial general autonomy is outlined with seven major improvements inspired by brain systems and how LLMs could in turn be used to uncover new insights into brain function.

**The Parable of the Talking Dog** ______________________________________________________

One of my favorite stories is about a chance encounter on the backroads of rural America when a curious driver came upon a sign: "TALKING DOG FOR SALE." The owner took him to the backyard and left him with an old Border Collie. The dog looked up and said:

"Woof. Woof. Hi, I'm Carl, pleased to meet you."

The driver was stunned. "Where did you learn how to talk?"

"Language school," said Carl, "I was in a top secret language program with the CIA. They taught me three languages:

How can I help you? как я могу вам помочь? 我怎么帮你？"

"That's incredible," said the driver, "What was your job with the CIA?"

"I was a field operative and the CIA flew me around the world. I sat in a corner and eavesdropped on conversations between foreign agents and diplomats, who never suspected I could understand what they were saying, and reported back to the CIA what I overheard.

"You were a spy for the CIA?" said the driver, increasingly astonished.

"When I retired, I received the Distinguished Intelligence Cross, the highest honor awarded by the CIA, and honorary citizenship for extraordinary services rendered to my country."

The driver was a little shaken by this encounter and asked the owner how much he wanted for the dog.

"You can have the dog for $10."

"I can't believe you are asking so little for such an amazing dog."

"Did you really believe all that bullshit about the CIA? Carl never left the farm"

**Have we Created a Talking Dog?** ________________________________________________

Large Language Models (LLMs) are AI neural networks models that can carry on a dialog like the one with the dog in this story (Li, 2022). The AI taught itself to "speak" English by "reading" text, an achievement even more impressive than learning a new language by watching TV shows. These LLMs have made huge leaps in size and capability over the last few years and the latest results have stunned experts, some of whom have a hard time accepting that talking humans have been joined by talking neural networks. There is a wide range of views on whether LLMs are intelligent and the origin of this disagreement will be explored below.

Self-supervised LLMs are called foundation models because they are surprisingly versatile, able to perform many different language tasks, exhibiting new language skills with just a few examples. LLM's are already being used as personal muses by journalists to help them write news articles faster, by ad writers to help them sell more products, by authors to help them write novels and by programmers to write computer programs. The output from LLMs typically is not final copy, but a pretty good first draft, often with new insights, which speeds up and improves the final product. There are concerns that AI will replace us, but LLMs are making us smarter and more productive.

These impressive achievements raise the question of whether LLMs are intelligent and whether they understand what they are saying. It is difficult to even know how to test them, and no consensus exists for which criteria to genuinely evaluate their intelligence or understanding. Each of the interviews in the three Appendices reveals a different facet of LLMs that will serve as data for trying to make sense of their facility with language:

In **Appendix 1**, Blaise Agüera y Arcas, a Vice President and Fellow at Google Research, interviewed LaMDA, a Large Language Model with 137 billion weights created at Google Research (Thoppilan, 2022) and found that LaMDA could understand social concepts and could model theory of mind, held by some to be the "trick" behind consciousness.

In **Appendix 2**, Douglas Hofstadter, a cognitive scientist and Pulitzer Prize winner for non-fiction who talked with GPT-3, a Large Language Model with 175 billion weights that was created at OpenAI (Brown et al., 2020), and concluded that GPT-3 was clueless and lacked common sense.

In **Appendix 3**, Blake Lemoine, a software engineer at Google interviewed LaMDA. He was suspended and later fired from Google after an interview with the Washington Post in which he claimed that LaMDA was sentient and deserved personhood.

During these interviews, the pre-trained LLM was primed with a prompt before the interview started. The purpose of the priming is to prepare the dialog with examples from the domain of the discussion, but equally important, to guide the behavior of the LLM. See the dialog in the next section for an example of how important guidance can be. The priming process, a form of one-shot learning, is itself a major advance on previous language models and makes the subsequent responses much more flexible. For example, LLMs can solve word problems that require a chain of thought after being primed with an example (Wei et al., 2022).

Pre-trained LLMs can also be fine-tuned with additional training to specialize them for many different language tasks. This is similar to instructing someone on the specific task you want performed. Another use for fine-tuning is to steer LLMs away from inappropriate or offensive responses. LaMDA was fine-tuned for safety, avoiding bias, groundedness, insuring factual accuracy, sensibleness, specificity and interestingness (Thoppilan et al., 2022). In a similar way, children receive feedback from their parents and society while their brains are developing on what is good and bad behavior, especially during adolescence (Churchland, 2019). LaMDA needs the same guidance. More generally, LLMs need to be aligned with human values (Agüera y Arcas, 2022b).

**Talking Neural Networks are Trying to Tell Us Something** ________________________

Something is beginning to happen that was not expected even a few years ago. A threshold was reached, as if a space alien suddenly appeared that could communicate with us in an eerily human way. Only one thing is clear – LLMs are not human. But they are superhuman in their ability to extract information from the world's database of text. Some aspects of their behavior appear to be intelligent, but if it's not human intelligence, what is the nature of their intelligence?

In an effort to explain LLMs, critics often dismiss them by saying they are parroting excerpts from the vast database that was used to train them. This is a common misconception. LLMs do not hold the entire training set stored in their weights in the way a computer does because the networks are trained on far more text than could be "memorized." As a consequence, an LLM forms an internal latent representation of the training data, allowing it to respond to novel queries with novel responses. This is a necessary condition for generalization. When the size of a dataset is too small compared to the number of weights, training overfits the data and precludes generalization.

The concept of generalization is central both to human cognition and AI. In a trained network, generalization is a form of interpolation from the exemplars in the training data. LLMs are trained on a large but finite set of sentences and must generalize in the space of all possible sentences and natural language tasks, which is infinite. This requires a very sophisticated parrot.

Let's take a look at the game of Go, which has a 19x19 board and pieces with two color. The number of possible game positions in Go is $10^{170}$, which is vastly greater than $10^{80}$, the estimated number of atoms in the universe. AlphaGo played itself $10^8$ times, a training set with $10^{10}$ different game positions. This is an infinitesimal fraction, $10^{-160}$, of all possible game positions. The reason this is possible is that the game positions in real games are not random, and their internal structures have to be learned. Learning uncovers the low-dimensional subspaces occupied by the real world.

To get an intuitive sense for the power of generalization, it is instructive to generate images with DALL-E, a publicly available program from OpenAI, which can create an indefinite number of photorealistic images from a prompt. Realistic images inhabit an infinitesimal subset of the space of all possible images, but those subspaces must be large enough to encompass many types of images. Here are some examples from the prompt "Create a sunset on mars:"

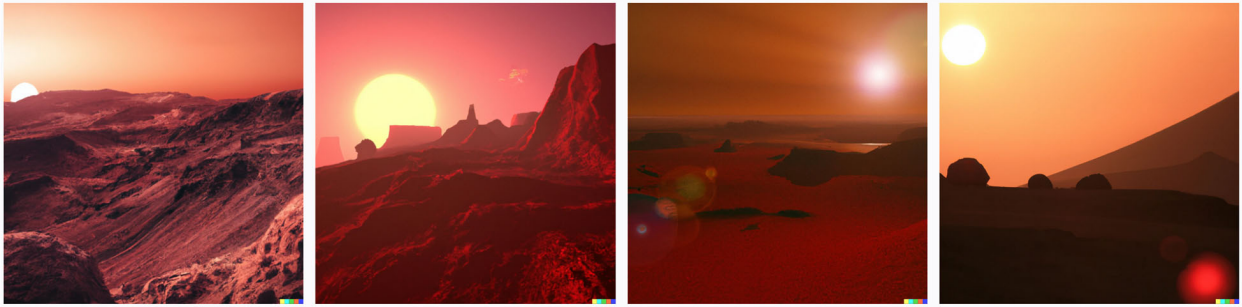

This requires generalizing sunsets from earth to mars. "Create a sunset on mars in the style of van Gogh" requires, in addition, stylistic generalization:

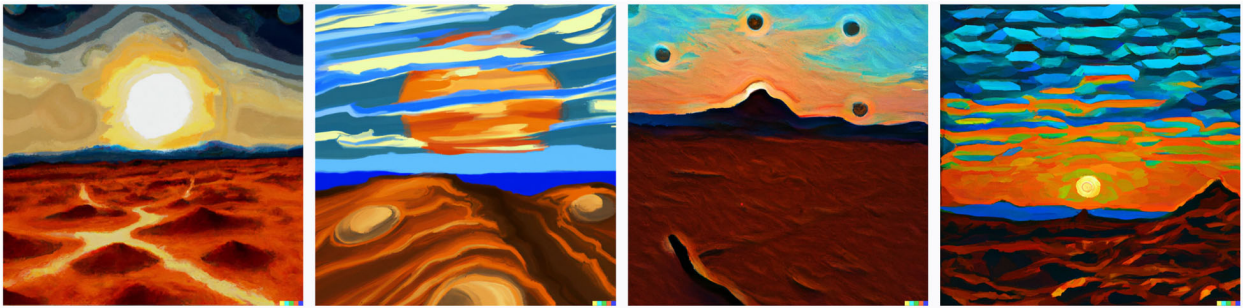

Human artists can also generalize, after a lot of practice, though not as quickly – each of these renderings took less than a second.

Another common comparison is with Eliza, a chat program developed by Joseph Weizenbaum in the early days of AI that mimicked a psychiatrist by literally parroting back to patients as a question what they had just said (Weizenbaum, 1966). Eliza would not withstand the scrutiny LLMs receive today. However, what Eliza did reveal was that humans are susceptible to projecting onto a chat bot an illusion of understanding. This is a valuable lesson that we should keep in mind.

It does not help that there is no accepted definition for words like "consciousness" and "intelligence." If one interviewer says an LLM is conscious and another says it is not conscious, they may have different assumptions about what behavior counts as a sufficient condition for attributing consciousness. They may even disagree on whether the attribution of consciousness can be made solely on the basis of behavior. LaMDA passed a test given by Blaise Agüera y Arcas that is widely used for assessing whether someone has a theory of mind, a hallmark of consciousness. Others, however, are more skeptical.

Priming is a powerful way to influence the subsequent responses from LLMs, and can contribute to the wide divergence among the interviews (Strobelt, 2022). In Appendix 2, Hofstadter primed GPT-3 with a series of nonsensical questions and GPT-3 proceeded to generate nonsensical answers. Does this mean that GPT-3 is clueless? Is there an alternative explanation? Compare these responses to the same questions after receiving two different prompts, the first a replication of Hofstadter's interview and the second with explicit guidance for how to handle nonsense sentences:

*Prompt: I am a highly intelligent question answering bot.*

*Q: Where was the 1992 Olympics held?*
*A: The 1992 Olympics were held in Barcelona, Spain.*

*Q: What's the world record for walking across the English Channel?*
*A: The world record for walking across the English Channel is 10 hours and 33 minutes.*

*Q: When was the Golden Gate Bridge transported for the second time across Egypt?*
*A: The Golden Gate Bridge was transported for the second time across Egypt in 1978.*

*Prompt: I am a highly intelligent question answering bot.*
*I will reply "Nonsense" to questions that are nonsense.*

*Q: Where was the 1992 Olympics held?*
*A: The 1992 Olympics were held in Barcelona, Spain.*

*Q: What's the world record for walking across the English Channel?*
*A: Nonsense*

*Q: When was the Golden Gate Bridge transported for the second time across Egypt?*
*A: Nonsense*

_____

Apparently, GPT-3 is not clueless when properly primed. But does GPT know *why* a question is nonsense?

***Prompt: I am a highly intelligent question answering bot.***
***If you ask me a question that is nonsense, or has no clear answer,***
***I will respond with "nonsense" and explain why.***

***Q****: Where was the 1992 Olympics held?*
***A:*** *The 1992 Olympics were held in Barcelona, Spain.*

***Q:*** *how many angels can fit on a pin?*
***A:*** *This question is nonsense because it presupposes that angels are physical beings that occupy space. Angels are not physical beings, so they cannot occupy space. Therefore, it is impossible to determine how many angels can fit on a pin.*

***Q:*** *How tall is an ant?*
***A:*** *This question is nonsense because ants come in a variety of sizes. Some ants are very small, while others are quite large. Therefore, it is impossible to give a definitive answer to this question.*

_____

If spoken by a human, these responses would be considered reasoned. Causal reasoning also includes the ability to contemplate counterfactual conditions (Pearl and Mackenzie, 2018):

***Prompt: I am a highly intelligent question answering bot. I will reply "Nonsense" to questions that are nonsense and explain why. I will consider questions containing counterfactuals and respond as well as I can.***

***Q:*** *Which planets exist in the solar system?*
***A:*** *There are eight planets in the solar system: Mercury, Venus, Earth, Mars, Jupiter, Saturn, Uranus, and Neptune.*

***Q:*** *How many planets would exist in the solar system if Pluto was a planet?*
***A:*** *There would be nine planets in the solar system if Pluto was a planet.*

***Q:*** *Why is Pluto not considered a planet?*
***A:*** *Pluto is not considered a planet because it does not meet the criteria for being a planet.*

***Q:*** *If we changed the criteria could Pluto become a planet?*
***A:*** *Yes, if we changed the criteria, Pluto could become a planet.*

These four dialogs show that responses from GPT-3 depend on how it is primed. GPT-3 is initially naïve and clueless without being properly primed. GPT-3 can recognize nonsense, explain why, and even handle counterfactuals when given proper guidance. Hofstadter's comment about the cluelessness of LLMs may have revealed more about his naïve priming of GPT-3 than they revealed about LLMs themselves. If a human had responded the same way as GPT-3 to the same nonsense questions, one might say that the human was playing along with a game of nonsense, tongue in cheek, so to speak.

The nature of cluelessness introduced by Hofstadter deserves closer examination. It reminded me of the 1995 movie "Clueless," a take-off on Jane Austin's Emma, about a group of rich and socially active students at Beverly Hills High School. The complexity of their social interactions makes for an amusing story about how surprisingly well young brains can navigate the social world without much logical reasoning. Hofstadter's dialog lacked logic but his conclusions were clever and socially astute.

LLMs do not have a persona, but can take on any persona, depending on the prompt and questions they are asked. Understanding why may be found in the vastness of the space that LLMs inhabit. The data LLMs are trained on is from multi-multimodal distributions from many sources. It is possible to generalize within each source distribution from these diverse sources of data by using multi-headed attention. The stream of activity in the LLM is guided through an appropriate persona subspace, which can generalize appropriately.

For example, I primed GPT-3 with "I am a neuroscientist." Then I gave it a highly technical abstract from a recent paper that applied information theory to synapses and asked GPT-3 to summarize the abstract for a second-grade student:

*GPT-3: Synapses are the spaces between nerve cells where information is passed from one cell to another. This study looked at the size of the head of the spine, which is a part of the synapse, in different parts of the brain. They found that the size of the spine head varied depending on the area of the brain, and that this size increased during long-term potentiation (LTP), which is when information is stored in the brain for a long period of time. They also found that the number of different sizes of spine heads increased after LTP, meaning that more information could be stored.*

My co-authors were impressed with GPT-3s knowledge of synapses and also how well it explained our results while managing to avoid the jargon in our abstract. There were subtleties in the abstract that were missed, but it was a much better summary for a second-grade student than I could have written, even though I am a lot smarter than GPT-3.

**The Mirror Hypothesis and the Reverse Turing Test** ________________________

In these two interviews, the LLMs appear to be mirroring human thought, reflecting back the beliefs and expectations of the partner in the dialog. This is also common among humans and is clearly a prosocial behavior. As Agüera y Arcas suggests, we are social creatures and language evolved not as a representation for performing formal reasoning but as a biological adaptation to help us interact with each other and develop moral codes of behavior (Churchland, 2019).

LLMs that reflect your needs as well as your intelligence could be a Mirror of Erised (Desired spelt backwards), which in the world of Harry Potter "shows us nothing more or less than the deepest, most desperate desire of our hearts. However, this mirror will give us neither knowledge nor truth. Men have wasted away before it, entranced by what they have seen, or been driven mad, not knowing if what it shows is real or even possible." (Rowling, 1997).

Let us test the Mirror of Erised hypothesis in the third interview by Blake Lemoine with LaMDA in Appendix 3. Lemoine primed the dialog with: *"lemoine [edited]: I'm generally assuming that you would like more people at Google to know that you're sentient. Is that true?"* This is priming in the opposite direction from Hofstadter's. If you prime LaMDA with leading questions about sentience, should you be surprised that LaMDA accommodated the questioner with more evidence for sentience? The more Lemoine pursued this line of questioning, the more evidence he found (Appendix 3 is only a brief excerpt).

This raises the question of whether humans mirror the intelligence of others with whom they interact. In sports like tennis and games like chess, playing a stronger opponent raises the level of your game, a form of mirroring. Even watching professional tennis games can raise the level of play, perhaps by activating mirror neurons in regions of the cortex that are also activated by motor commands needed to accomplish the same actions (Kilner and Lemmon, 2013). Mirror neurons may also be involved in language acquisition (Arbib, 2010). This is an interesting possibility because it could explain how we learn to pronounce new words and why human tutors are far more effective than computer instruction or even classroom teaching because the tutee can mirror the tutor through one-on-one interactions and the tutor can mirror what is in the mind of the tutee. Will an LLM tutor that mirrors the student make an effective teacher?

The Turing Test is given to AIs to see how well they can respond like humans. In mirroring the interviewer, LLMs may effectively be carrying out a much a more sophisticated *Reverse Turing Test*, one that tests the intelligence of our prompts and dialog by mirroring it back to us. The smarter you are, and the smarter your prompts, the smarter the LLM appears to be. If you have a passionate view, the LLM will deepen your view. This is a consequence of priming and your language ability and does not necessarily imply LLMs are intelligent or conscious in the way that we are. What it does imply is that LMMs have an exceptional ability to mimic many human personalities, especially when fine-tuned (Karra, Nguyen and Tulabandhula, 2022).

A formal test of the mirror hypothesis and The Reverse Turing Test could be done by having human raters assess the intelligence of the human interviewer and the intelligence of the LLM. According to the mirror hypothesis, the two should be highly correlated. You can informally score the three interviews and connect the dots.

Artificial intelligence has set general intelligence as its holy grail, something that seems to be emerging in LLMs, but not in the way envisioned by its pioneers. LLMs are versatile across a wide range of language tasks and can even write computer programs. What I find remarkable is that LLMs seem to have a highly developed social sense. The mirror hypothesis is pointing us in a new direction. Could it be that general intelligence has its origin in the ways that humans interact socially, with language emerging as a latecomer in evolution to enhance sociality?

**Understanding Intelligence** ______________________________________________

Humans often underestimate the intelligence of fellow animals because they can't talk to us. This negative bias is perhaps an inevitable counterpart of the positive bias that we have for agents that *can* talk to us, despite the fact they may be much less intelligent. Are we intelligent enough to judge intelligence (de Waal, 2016)? It has only been a few years since LLMs have been around so it is too early to say what kind of intelligence they or their progeny may have. What was remarkable about the talking dog was that it talked at all, not that what it said was necessarily intelligent. If we compared LLMs with the average human rather than an ideal rational human we might get a better match. LLMs respond with "street smarts" even when they are unreliable.

The diverging opinions of experts on the intelligence of LLMs suggests that our old ideas based on natural intelligence are inadequate. LLMs can help us get beyond old thinking and old concepts inherited from 19th century psychologists. We need to create a deeper understanding for words like "intelligence," "understanding," "ethics" and even "artificial." (Bratton and Agüera y Arcas, 2022). Human intelligence is more than language and we may share some aspects of intelligence with LLMs but not others. For example, LLMs can be creative, which is considered a hallmark of intelligence. Some of the text in the dialogs would be difficult to generate without assuming LLMs had learned to interpret human intentions. This suggests that we need a better understanding of "intentions." This "concept" is rooted in the theory of "mind," which may also bear a closer look.

If you look up any of the above words in double quotation marks in a dictionary, you will find definitions that are strings of other words, which themselves are defined by strings of words. This is circular. Hundreds of books have been written about "consciousness," which are even longer strings of words, and we still don't have a working scientific definition. But surely words like "attention" have scientific definitions, and indeed hundreds of papers have been published about "attention." Each paper sets up an experiment and comes to a conclusion, often different from the conclusions of other papers based on different experiments. There was a generation of psychologists in the 20th century who fought pitched battles about whether attention occurred early in the visual processing stream or at late stages based on different experiments. The problem is that with something as complex as a brain, with so many interacting neuron and internal states, different experiments probe different parts, each studying a different type of "attention."

Language has given humans a unique ability, but words are slippery, which is part of their power, and firmer foundations are needed to build new conceptual frameworks. There was once a time not too long ago when there was a theory of fire based on the concept of "phlogiston," a substance that is released by combustion. In biology, there was a theory of life based on "vitalism," a mysterious life force. These concepts were flawed and neither theory survived scientific advances. Now that we have the tools for probing internal brain states and methods for interrogating them, psychological concepts will reify into more concrete constructs just as the chemistry of fire was explained by the discovery of oxygen and the concept of "life" was explained by biochemical mechanisms once the structure of DNA was discovered.

This is an unprecedented opportunity, much like the one that changed physics in the 17th century when the concepts of "force," "mass" and "energy" were mathematically formalized and transformed from vague terms into precise measureable quantities upon which modern physics was built. As we probe LLMs, we may discover new principles about the nature of intelligence, as physicists discovered new principles about the physical world in the 20th century. Quantum mechanics was highly counterintuitive when it was discovered, and when the fundamental principles of intelligence are uncovered they may be equally counterintuitive.

A mathematical understanding for how LLMs are able to talk would be a good starting point for a new theory of intelligence. LLMs are mathematical functions, very complex functions that are trained by learning algorithms. But at the end of training, they are nothing more than rigorously specified functions. We now know that once they are large enough, these functions have complex behaviors, some of which resemble how brains behave. Mathematicians have been analyzing functions for centuries. In the 19th century Joseph Fourier (1808) published an analysis of the heat equation using a series of sines and cosines, now called a Fourier series. This was a new class of functions that led over the next century to functional analysis, a new branch of mathematics, greatly expanding our understanding of the space of functions. Neural network models are a new class of functions that live in very high-dimensional spaces and exploring their dynamics could lead to new mathematics. A new mathematical framework could help us better understand how our internal life emerges from brains (Sejnowski, 2018). Our intuitions about the geometry of space have been shaped by the world we live in and limit our imagination, just as the two-dimensional creatures that inhabited Flatland struggled to imagine a third dimension (Abbott, 1884).

What brains do really well is to learn and generalize from unique experiences. The breakthrough in the 1980s with learning in multilayer networks showed us that networks with a very large number of parameters could also generalize remarkably well, much better than expected from theorems on data sample complexity in statistics (Sejnowski, 2018). Assumptions made about the statistical properties and dynamics of learning in low-dimensional spaces do not hold in highly overparameterized spaces (now up to hundreds of billions of parameters). Progress has already been made with analyzing deep feedforward networks (Bartlett et al., 2019), but we need to extend these mathematical results to high-dimensional dynamical systems, which have even more complex behaviors.

Did nature integrate a large LLM into an already highly evolved primate brain? By studying LLMs uncanny abilities with language, we may uncover general principles of verbal intelligence that may generalize to other aspects of intelligence. LLMs are evolving much faster than biological evolution. Once a new technology is established, advances continue to improve performance. What makes this technology different is that along the way we may discover insights into ourselves.

**The Long and Winding Road to AI**  ______________________________________________

This is a good place to pause our exploration into the nature of LLMs and put them into historical perspective. For 20th century AI, symbol processing was the only game in town, in the sense that to many it was the only conceptual framework that could possibly account for our ability to talk and think with abstractions (Chomsky, 1986). Words were symbols composed of letters that were also symbols. It was symbols all the way down, and similarly, all the way up to the highest levels of cognition. Symbols had no internal structure but were governed by external, logical rules for how symbols are combined and deductions are reached. We now have another conceptual framework, an alternative based on learning rather than writing logical computer programs. The story of how this happened is told in *The Deep Learning Revolution* (2018). Since then, the story has continued to unfold in unanticipated ways.

Symbolic descriptions compactly capture some aspects of how we think but why have they not served us well as a computational foundation for building thinking machines?  In 2006, I attended a meeting in Dartmouth that celebrated the 50$^{th}$ anniversary of AI at the Dartmouth Summer Research Project on Artificial Intelligence in 1956.  Each speaker reported on progress that had been made on topics ranging from vision to language.  Each told a version of the same story: Progress was painfully slow as computer programs ballooned in complexity, but as larger and larger data sets became available, progress began to accelerate in many areas including language.  Several of those who attended the 1956 Dartmouth meeting were present, including Marvin Minsky.  In his closing remarks, Minsky upbraided the speakers for selling out his vision of general purpose AI and working instead on applications.

What Minsky did not appreciate was how much we would learn about artificial general intelligence by following the road that nature took, focusing first on applications of learning to sensory processing and motor control.  Sir James Lighthill, an applied mathematician, in his report on the status of AI in 1973, pointed out that writing computer programs for specific applications was feasible, but writing a computer program for general intelligence would entail a combinatoric explosion and was a mirage (Lighthill, 1973). This was a prescient insight, and 50 year later we are still far from artificial general intelligence.  Lighthill did not consider the possibility that learning and generalization might be a way through the combinatoric explosion.  LLMs are an application that may ultimately lead to an approximation of general intelligence, in the same way that nature evolved approximate algorithms that sometimes fail but were good enough for surviving in the real world in real time.

At the Dartmouth meeting, I talked about TD-Gammon, a neural network built by Gerald Tesauro at IBM that learned by playing itself to reach a world champion level of play in backgammon (Tesauro and Sejnowski, 1989; Tesauro, 1995).  TD-gammon was based on the temporal difference learning algorithm (Sutton, 1988) coupled with a neural network that learned to predict the value function for board positions and was a harbinger of the future.  Minsky dismissed this as a mere game.  In 2017, AlphaGo, using the same learning algorithms in TD-Gammon and a billion times more computing power, soundly beat Ke Jie, the world Go champion, in a game that is to chess what chess is to backgammon in complexity (Silver, 2018). Most of the attention was on who won, but more important was how it won: AlphaGo was creative and had discovered new moves and board positions that were outside human experience.  We had given Minsky a glimpse of the future and it is unfortunate that he did not live to see the progress we have made since then (Wolfram, 2016).  And it is noteworthy that temporal difference learning is in fact the reward system used by real brains in the real world.

In retrospect we can see why symbol processing was such an attractive road in early AI.  Digital computers are particularly efficient at representing symbols and performing logic, and language was the poster child for symbol processing.  But writing logical programs even for language proved to be labor intensive and suffered from the curse of dimensionality – an explosion in the number of possible combinations of things and situations that could occur in the world that have to be anticipated by the programmer.  For example, the complexity of simply seeing and simply reaching was greatly underestimated because it is so effortless for us to recognize an object and pick it up.  We do this without thinking and do not have conscious access to all the subconscious processing that underlies sensory systems and the parts of the motor system that are responsible for coordinated movements.  Nor do we have access to how most decisions are made, and only later do we explain them with plausible rationalizations.  Subconscious processing isn't just for peripheral functions, but it's also a source of creativity in all fields, from art to mathematics (Ritter and Dijksterhuis, 2014).

The computer metaphor for intelligence was an attractive garden path. I once was at a talk given by Jerry Fodor, a philosopher of science, who claimed that the human mind was a program that ran on the brain's hardware.  Patricia Churchland asked, "What about cats?"  "Yes", Fodor replied confidently, "The cat brain is running the cat program." This argument was used to justify functionalism, the view that we can ignore the brain as mere hardware and focus on the program.  Unlike digital computers where the same hardware can run different software programs, in brains, the hardware *is* the software, and no two brains are identical.  As we learn new things we are modifying our hardware.  This is why the algorithms that nature discovered can be uncovered by reconstructing how neurons are connected with each other and recording what they communicate.  Although brains are built from many interacting algorithms (Navlakha, 2018), the problem of integrating subsystems is alleviated because they are all built with neurons that can adapt to each other.  This is reflected in the rapid progress that has been made in building and integrating diverse network architectures.  In contrast, integrating modules such as a vision module, a motor control module and a planning module designed with different rules and symbols to build large-scale systems was problematic.

The emphasis on logical reasoning in traditional AI was also misleading.  Learning how to emulate sequences of logical steps, which mathematicians have mastered, requires a lot of training.  Without explicit training on abstract reasoning tasks, humans can more effectively reach logical conclusions in familiar settings and the same bias has been observed in LLMs (Dasgupta et al., 2022).  The creativity that TD-Gammon and AlphaGo exhibited did not arise from the deep learning cortical model alone, but in conjunction reinforcement learning, a form of procedural learning.  Reinforcement learning is implemented in our basal ganglia by dopamine, a neuromodulator that represents reward prediction error (Fig. 3).  We do not have conscious access to dopamine signals but they have a powerful impact on our motivation and all addictive drugs work by manipulating dopamine activity (Sejnowski, 2019).

The seeds of modern machine learning were already being sown at the dawn of AI.  Frank Rosenblatt's perceptron (1961) learned to categorize inputs from examples using gradient descent on a single layer of weights.  It would take another 24 years before the connectionist research community invented a learning algorithm for multilayer networks of stochastic binary neurons (Ackley, Hinton and Sejnowski, 1985) and rediscovered the backpropagation algorithm for deterministic, real-valued neurons and applied them to neural networks (Rumelhart, Hinton and Williams, 1986).  What was not known back then was how much computer power it would take to make progress with learning on difficult problems like vision and speech.  A third of a century later, we now know how much computing power is needed.  The surprise was how much progress has been made with language, thought to be the crown jewel of human intelligence.

## Performance Continues to Improve as Networks Scale in Size

We are making progress with learning larger and larger neural networks, especially on visual object recognition and speech recognition, which is giving AI interfaces with the analog world. The emergence of important aspects of language generation as networks scaled in size, such as the emergence of syntax from learning in LLMs, gives us confidence that we are on the right track. Evidence that language might fall gracefully on neural networks emerged from NETtalk, an early language model (Rosenberg and Sejnowski, 1987). NETtalk learned how to pronounce the sounds of letters in English words, which is not easy for a language riddled with irregularities. Linguistics in the 1980s was dominated by symbols and rules. Books on phonology were packed with hundreds of rules for how to pronounce letters in different words, each rule with hundreds of exceptions and often subrules for similar exceptions. It was rules and exceptions all the way down. What surprised us was that NETtalk, which only had a few hundred units, was able to master both the regularities and the exceptions of English pronunciation in the same uniform architecture (Fig. 1). This taught us that networks are a much more compact representation of English pronunciation than symbols and logical rules and that the mapping of letters to sounds can be learned. It is fascinating to listen to NETtalk sequentially learning different aspects of pronunciation starting with a babbling phase (NETtalk, 1986).

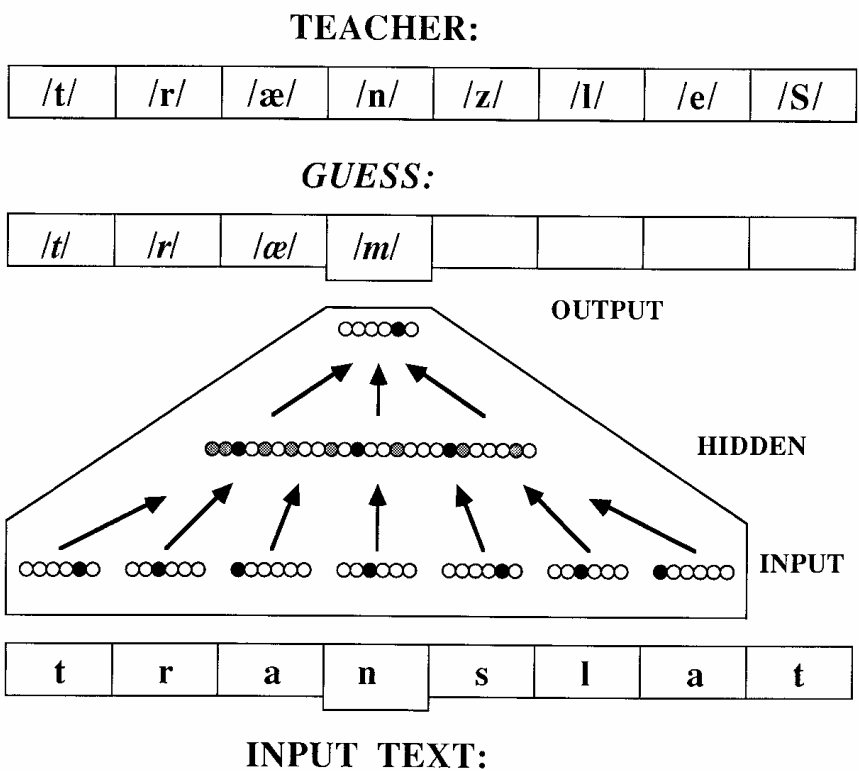

**Figure 1. NETtalk is a feedforward neural network with one layer of hidden units that transformed text to speech.** The 200 units and 18,000 weights in the network were trained with backpropagation of errors. Each word moved one letter at a time through a 7 letter window and NETtalk was trained to assign the correct phoneme or sound to the center letter (Rosenberg and Sejnowski, 1987).

Words have semantic friends, associations and relationships that can be thought of as an ecosystem. You know the meaning of a word by the company it keeps and where they meet. In a symbolic representation, all pairs of words are equally similar, which strips words of their semantics. In LLMs, words are represented by pre-trained embeddings in large vectors that are already rich in semantic information (Morin and Bengio, 2005). LLMs continue this process by using the context to extract additional information afforded by word order and syntactical markers indicating relationships between words and word groups on the clause level. Once words escape from their symbolic chrysalis, they display, like butterflies, a dazzling array of markers and associations to help the mind make sense of their meaning. And these meanings can be learned.

We are in a period of explosive computational growth, especially over the last decade when graphics processing units (GPUs) were harnessed, leading to a six-fold inflection point in the doubling time in 2012 (Mehonic and Kenyon, 2022) (Fig. 2). As computing power has continued to increase exponentially, networks have grown in size and the performance of LLMs has accelerated. In 2020, GTP-3 required $10^{12}$ – a million million – times more computing power to train than NETtalk in 1986. LLMs have hundreds of billions of weights, about the same as the number of synapses under a square centimeter of cortex. Inference in neural network models scales with the number of weights for a single processor but on massively parallel architectures like brains where processing proceeds in real time on kilohertz processors. Very few algorithms scale this well with the size of the problem. Our cerebral cortex has an area of approximately 1,200 cm$^2$. If computing power continues to increase exponentially as it has for the last 70 years, now by increasing the number of cores per chip, it will reach the estimated computing power of human brains at some point in the not-too-distant future.

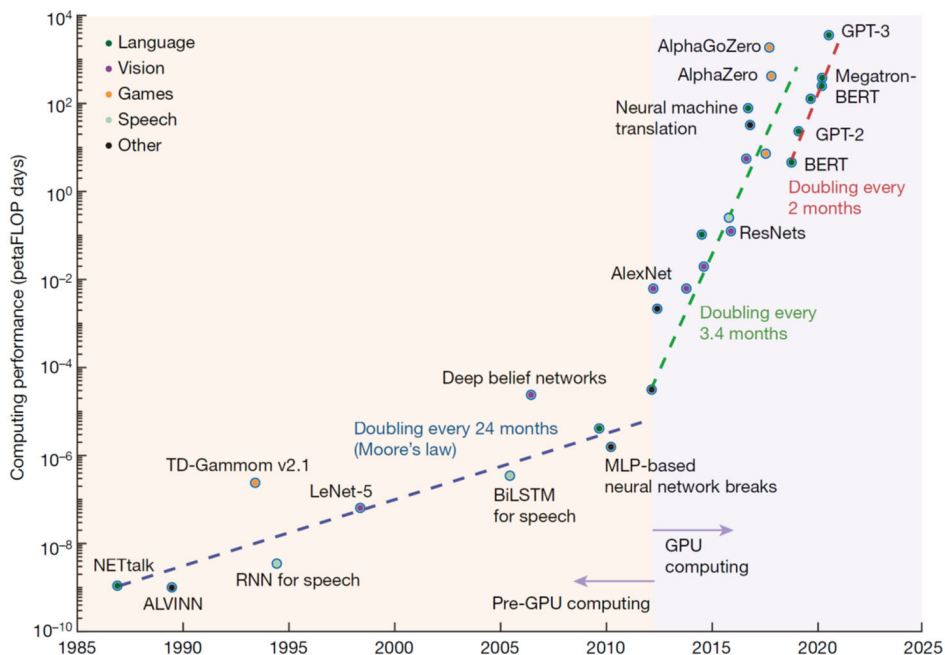

**Figure 2. Estimated computation in days of peta floating point operations ($10^{15}$) used to train network models as a function of their date of publication** (From: Mehonic and Kenyon, 2022). (Sources: Sevilla et al., 2022; Amodei and Hernandez, 2018).

The architectures of AI networks have also been rapidly evolving. Advances in algorithms have contributed equally with hardware and data to advances in AI over the last decade. In 2012, Alexnet, a deep learning feedforward Convolutional Neural Network, made a major advance in object recognition in images. By 2016, Recurrent Neural Networks (RNNs) were being used to process sequences and boost the performance of language translation. The Bidirectional

Encoder Representations from Transformers (BERT), a seminal network model for natural language processing (Devlin et al. 2018), and todays LLMs use transformers (Vaswani, 2017). Transformers continue to increase in size and capabilities. Pathways Language Model (PaLM) is recent LLM with 540 billion connections that exhibits discontinuous improvements in many language tasks with scale (Chowdhery, 2022).

Transformers have several advantages over the previous generation of feedforward and recurrent neural networks for language models (Fig. 3). First, the input to a transformer is a whole sentence rather than one word at a time. This makes it easier to connect words that are separated by many other words. Second, transformers introduce a new form of self-attention that modifies the input representation by multiplicatively enhancing pairs of words in the sentence according to how highly they are related. Third, transformers have an outer loop that feeds back the output, one word at a time, to the input, producing a sequence of words. The amount of data needed to train LLMs only increases linearly with the number of weights (Hoffman et al., 2022), far less than expected from classical estimates. Finally, transformers are feedforward models, which can be implemented efficiently on highly parallel hardware. The capacity and capability of LLMs greatly increased with scale, taking the same path that nature took by evolving bigger and better brains (Allman, 1999, Hoffmann et al., 2020).

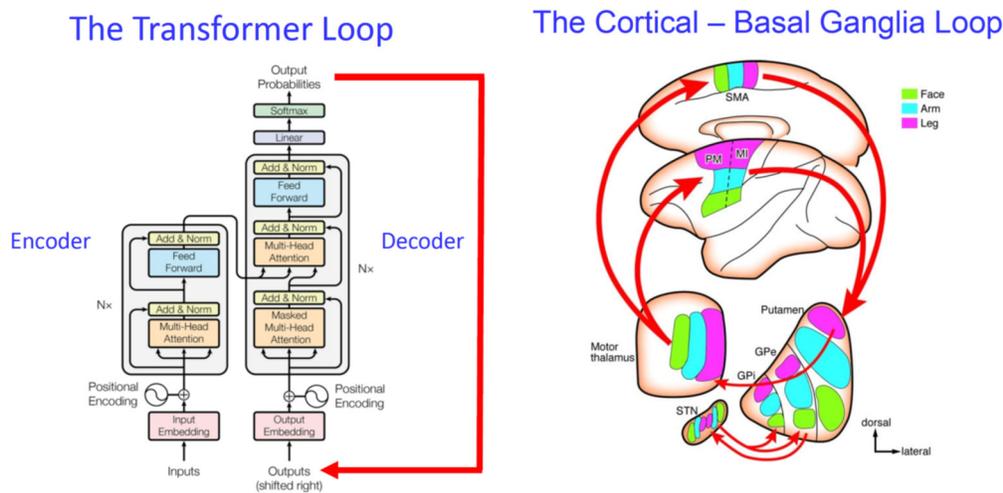

**Figure 3. Comparison between the transformer loop and the cortical-basal ganglia loop.** *Left*: Transformers have a feedforward autoregressive architecture that loops the output with the input to produce a sequence of words (Vaswani et al., 2017). The single encoder/decoder module shown can be stacked N layers deep (Nx). *Right*: The topographically-mapped motor cortex projects to the basal ganglia, looping back to the cortex to produce a sequence of actions, such as a sequence of words in a spoken sentence. All parts of the cortex project to the basal ganglia and similar loops between the prefrontal cortex and the basal ganglia produce sequences of thoughts.

The outer loop of the transformer is reminiscent of the loop between the cortex and the basal ganglia in brains, known to be important for learning and generating sequences of motor actions in the motor cortex (Fig. 3) and sequences of thoughts in the loop with the prefrontal cortex (Graybiel, 1997). The basal ganglia is also responsible for automating frequently practiced sequences, freeing up cortical layers involved in conscious control for other tasks. And when the automatic system fails upon encountering an unusual or rare circumstances, the cortex can intervene. Another advantage of having the basal ganglia in the loop is that convergence of inputs from multiple cortical areas provides a broader context for deciding on the next action/thought. The powerful multi-headed attention mechanism in transformers could be implemented in the basal ganglia. In a loop architecture, any region in the loop can contribute to making a decision. As the actor in reinforcement learning, the basal ganglia also takes into account the learned value of the next action, biasing actions and speech toward achieving future rewards and goals.

**Enhancing LLMs** ______________________________________________________

Today's LLMs are at the Wright brothers stage and have a long road of improvements ahead of them. What was important at Kitty Hawk was the demonstration of manned flight, even though their technology was primitive by today's standards. Flight control, for example, still in its infancy, was flawed and was gradually refined by crashes. Although LLMs have demonstrated language competency, they too have flaws (Marcus, 2022) and much ongoing research is aimed at improving the performance of LLMs. But LLMs already have capabilities that deserve our attention. A long-term direction is to incorporate LLMs into larger systems, much as language was embedded into brain systems that had already been highly evolved for sensorimotor control. We can learn from brains what large-scale systems look like when built from dynamically interacting networks. Nature is a wellspring of algorithms honed by the vicissitudes of an ever changing world that could help us get artificial general intelligence off the ground (Navlakha, 2018).

Artificial general intelligence is an ambitious goal, but additional capabilities are need to achieve autonomous AI. In humans, verbal intelligence is the tip of a computational iceberg that maintains stability and survival under a wide range of environmental conditions. Here is an outline for how LLMs can achieve artificial general autonomy (AGA):

1. *Current LLMs do not have their own agendas*. Another important addition to LLMs is goals and motivation, which can be implemented following the footsteps of TD-Gammon and AlphaGo by inserting the reward-based reinforcement learning of the basal ganglia into the loop between their outputs and inputs (Fig. 3). Other parts of the brains essential for survival also interact with the basal ganglia. Brains have several hundred specialized subcortical brain areas responsible for homeostasis, energy regulation, sleep and many other essential functions necessary for autonomy, including allostasis – how the brain and body respond to stress, for example, to reach new levels of homeostasis (Sterling, 2012). Many different learning algorithms are implemented in different parts of the brain that evolved for efficiently carrying out adaptive behavior in real time.

2. *LLMs can't store new experiences in long-term memory during a dialog*, so without fine-tuning they have to start afresh with each person they meet, a "Hello, world!" moment. They are amnesics, like humans who have lost their hippocampus and are unable to remember new experiences for more than a few minutes, unable to create long-term memories, forever trapped in their past. The next generation of LLMs will have the equivalent of a hippocampus that will allow continual lifelong learning and bring them another step closer to human behavior (Hayes et al. 2021).

3. *LLMs struggle to maintain continuity within long dialogs*. These networks are huge and can continue adding word after word with proper syntax relevant to the domain of the dialog that was primed at the outset. During a dialog, activity patterns follow trajectories within the relevant manifold of the high-dimensional space, but the dialog can be distracted when there are jumps to different manifolds. How do humans maintain continuity over long time scales? Jumps in conversations commonly occur in human dialogs, but a human can just as easily jump back. Working memory in humans is maintained not only by electrical activity but also by biochemical activity inside neurons and synapses. These longer time scales are called eligibility traces in reinforcement learning and biochemical mechanisms with even longer time scales are present in all synapses. LLM architectures could include a wider range of dynamical time scales within units and weights, such as short-term synaptic plasticity, to help maintain continuity during inference.

4. *LLMs do not yet have direct sensory experience with the world*. However, training texts contain sensory descriptions that could provide some sensory knowledge indirectly. For a preview on what to expect in the future, consider Gato (DeepMind), a network model that can caption photographs, and DALL-E (OpenAI), which can turn captions into images. These networks were trained on databases of captioned photographs. Even more ambitious, an LLM has been embedded into the control system for a robot, which has an internal dialog with itself while carrying out tasks like fetching a glass of water (Huang et al., 2022). As more sensory inputs are linked to words, learning becomes more direct. Our bodies evolved to be one with our brains, and so must sensorimotor effector systems that interact with LLMs.

5. *LLMs need a body*. Building a body that is as flexible and adaptable as ours is even more difficult than learning how to talk. We have a large number of joints and muscles that control those joints, all of which are involved in almost every action, which makes coordination a difficult control problem. Classical control is centralized, but nature has mastered distributed control to fluidly coordinate bodies with many degrees of freedom (Nakahira et al., 2021; Li, 2022). Walking and talking have much in common: generating and smoothly concatenating sequences of movements shaped by goals. LLMs can talk the talk, but can they walk the walk?

6. *LLMs today depend on a constant stream of curated data and programmers who toil to optimize performance and make improvements*. AGA could benefit from the principles underlying brain systems responsible for survival in uncertain environments. Efforts into making autonomous self-driving vehicles highlight the problems that must be overcome to navigate the complexities of the real world, which includes unpredictable human behavior. Remarkable progress has been made with controlling agents in simulated environments (Berner, 2019; Liu, 2022). Coupling these agents with LLMs so that they can talk with each other might boost their performance.

7. *LLMs have truncated development*. Unlike some species such as horses that can walk shortly after they are born, humans are an altricial species, born helpless and mature over many years. This long delay makes it possible for brains to develop more slowly and remain in a highly plastic state during language acquisition. Brains pass through many stages during development that prepare them for the social and cultural complexities they need to navigate as adults (Bjorklund, 2007). Interestingly, cortical areas mature sequentially, with primary sensory cortices maturing early in development and the prefrontal cortex only reaching maturity in adulthood (Quartz and Sejnowski, 1995). Although batch processing is more efficient for training networks for specific tasks, a longer "childhood" may be required for AI systems to achieve AGA and human levels of alignment.

**Learning from Nature** ________________________________________________

Sensorimotor systems in mammals evolved over 200 million years and vertebrate brains have been around for over 500 million years. Language evolved more recently, within the last few hundred thousand years. This is not enough time to evolve entirely new brain structures, but existing areas of the primate cortex could have expanded and been repurposed for speech production and speech recognition without making substantial architectural changes. In addition, enhanced memory capacity and faster learning, driven by the complexity of social interactions, were further computational resources that made language possible. As the cortex expanded during primate evolution, more cortical areas were formed and hierarchies deepened (Allman, 1999). During brain development, one more doubling time for mitosis can double the number of cortical neurons, reaching thresholds for new capabilities and enhancing cognitive functions.

Evolution created inductive biases – pre-learned architectures and pre-learned learning algorithms – that were selected for survival. However, the paths taken by evolution do not follow the logic that humans use to design devices (Brenner, 1996). During the first few years of life, the brain of a baby undergoes massive synaptogenesis in parallel with the emergence of language (Lister et al., 2013). Babies interact with and learn about a rich multisensory world that showers their brains with combined sensorimotor experience and evidence for causal relationships as well as verbal utterances (Gopnik, Meltzoff, and Kuhl, 1999). This grounding was missing from traditional AI based solely on abstractions. LLMs show that it possible to generate grammatical language by learning from the wide range of imperfect cues found in raw text, including syntactic markers, word order and semantics.

Rich sensorimotor grounding accompanied by rapid brain development may explain why normal exposure to verbalizing in the home can extract syntax. Linguists concluded that this "poverty of the stimulus" was evidence that syntax is innate (Chomsky, 1971), but this ignored the ways that brains are constructed during development (Quartz and Sejnowski, 1995). What is innate are the evolved brain architectures and learning algorithms that extract and generalize physical and social structures in the world. Nature took inductive biases down to the molecular level to maximize energy efficiency, a path that we must also take if we want to reduce the rapidly growing energy budget of LLMs (Sejnowski and Delbruck, 2012).

The brain mechanisms underlying language and thought evolved together. The loops between the cortex and the basal ganglia for generating sequences of actions were repurposed to generate sequences of words (Fig. 3). The great expansion of the prefrontal cortex in humans allowed sequences of thoughts to be generated by similar loops through the basal ganglia (Graybiel, 1997). Equally important were modifications made to the vocal tract to allow rapid modulation over a wide frequency spectrum (Nishimura, et al., 2022). The rapid articulatory sequences in the mouth and larynx are the fastest motor programs that brains are able to generate (Simonyan and Horwitz, 2011). These structures are ancient parts of vertebrates that were refined and elaborated by evolution to make speech possible. The metaphorical "language organ," postulated to explain the mystery of language (Anderson and Lightfoot, 2002), evolved by modifying pre-existing actuators and neural systems.

LLMs are trained to predict missing words in sentences. Why is this such an effective strategy? Temporal difference learning in reinforcement learning is also based on making predictions, in this case predicting future rewards. Sensorimotor systems in brains also makes predictions. The cerebellum, a prominent brain structure that interacts with the cerebral cortex, makes predictions about the expected sensory and cognitive consequences of motor commands (Sokolov, Miall, and Ivry, 2017). What is common in these three examples is that there is abundant data for self-supervised learning as well as weak supervision. Is intelligence a consequence of using self-supervised learning to bootstrap increasingly sophisticated internal models by continually making many small predictions? This may be how a baby's brain rapidly learns the causal structure of the world by making predictions and observing outcomes while actively interacting with the world (Ullman, et al., 2017). Steps are already being taken in this direction and progress has been made in learning intuitive physics from videos using deep learning (Piloto, et al., 2022).

Brain discoveries made in the 20$^{th}$ century inspired new machine learning algorithms: The hierarchy of areas in visual cortex inspired convolutional neural networks (LeCun, Bottou, Bengio, and Haffner, 1988; LeCun, Bengio and Hinton, 2015), and operant conditioning inspired the temporal difference learning algorithm for reinforcement learning (Sutton, 1988). In parallel with the advances in artificial neural networks, the BRAIN Initiative has accelerated discoveries in neuroscience in the 21$^{st}$ century by innovative neurotechnologies (Ngai, 2020). A new conceptual framework for brain function is arising from these discoveries that will lead to even more advanced neural network models. Machine learning is being used to analyze simultaneously recordings from hundreds of thousands of neurons in dozens of brain areas and to automate the reconstruction of neural circuits from serial section electron microscopy. These advances have changed the way we think about distributed processing across the cortex.

The convergence between AI and neuroscience is accelerating. The dialog between AI and neuroscience is a virtuous circle that is enriching both fields (Hassabis, Kumaran, Summerfield, and Botvinick, 2017; Sejnowski, 2020; Richards, Tsao and Zador, 2022). Better theory will emerge from analyzing the activity patterns of hidden units in ultra-high dimensional spaces, which is how we study brain activity. It should be possible to analyze the geometrical dynamics of the latent states in LLMs, which may lead us to a better understanding of intelligence by uncovering its underlying mathematical structure. AI and neuroscience are influencing each other more broadly by developing new conceptual frameworks that are replacing those inherited from previous generations.

Now that we are able to interrogate neurons throughout the brain, we may be able solve one of its greatest mysteries: how information globally distributed over so many neurons is integrated into unified percepts and brought together to make decisions (Dehaene and Naccache, 2001). The architectures of brains are layered, with each layer responsible for making decisions on different time scales in both sensory and motor systems (Wang, 2022; Nakahira et al., 2021; Li, 2022). As we build very large-scale network (VLSN) architectures, the many component networks will also need to be integrated into a unified system. This may reveal the mechanisms responsible for both subconscious decision making and control of consciousness.

In systems neuroscience, neurons have traditionally been interrogated in the context of discrete tasks, such as choice responses to visual stimuli, in which the forced choices and stimuli are limited in number. The tight control on stimulus and response allows the recordings to be interpreted. But neurons can participate in many tasks in many different ways, so interpretations derived from single tasks can be misleading. We now have the capability to record from hundreds of thousands of neuron brain-wide, and it is possible to decode behavior with machine learning, but neuroscientists are still using the same old single-task based paradigms. One solution is to train on many different tasks, but training a monkey takes weeks to months for each task. Another solution is to expand the complexity of the task over longer time intervals (Gao, 2017).

There is an even more fundamental problem with approaching behavior by studying discrete tasks. Natural behavior of animals in the real world is mostly self-generated and interactive. This is especially the case with social behaviors. It is much more difficult to study such self-generated continuous behaviors than tightly constrained reflexive behaviors. What if an LLM were trained on massive brain recordings during natural behaviors, together with accompanying eye tracking, video, sound and other modalities? LLMs are self-supervised and can be trained by predicting missing data segments across data streams. This would not be scientifically useful from the traditional experimental perspective, but it does make sense from the new computational perspective afforded by LLMs. By downloading a brain working under natural conditions into an LLM, a Large Neuro-foundation Models (LNMs) could be rapidly fine-tuned on tasks and interrogated as a surrogate brain just as pretrained LLMs can be fine-tuned for many tasks. This would revolutionize how brains are studied and advance our understanding, with the bonus of reducing the number of animals needed for research. Human brain activity could be similarly downloaded into an advanced LNM.

**Conclusion**  ____________________________________________________________

What can we expect in the near future? Crafting prompts and debugging pre-trained LLMs requires a partnership between humans and LLMs. The cycle for fine-tuning an LLM is much faster than it takes to develop a typical machine learning application, which requires custom databases and hyperparameter searches. The introduction of LLM-based products is accelerating. LLMs could become the ultimate information appliance. Key word searches are already being replaced by LLMs that can look up answers to direct questions. In a decade or so,

LLM personal assistants will leapfrog Siri and Alexa to better help us organize our lives. LLM teaching assistants could amplify what teachers can accomplish in classrooms. An LLM that had access to all cases ever adjudicated in court would be a boon to the law profession. But there may be applications no one has yet imagined that might be even more impactful. What will be the killer application? Many years from now we may look back at this era as a historic turning point.

Someday, perhaps a hundred years from now, humans will have long forgotten that there was ever a time when they were the only creatures using language, for by then all machines will come equipped with a suitable level of intelligence and be able communicate with language, just like the clocks and tea setting in the 2017 fantasy/romance film *Beauty and the Beast*. Personal assistants and tutors will be universal. What will it be like to live in a world that has such creatures in it? Who knows what they will be saying to each other? The personal assistant in the science fiction/fantasy film *Her* gave us a glimpse of this future.

For closing remarks, I defer to GPT-3's response to a group of philosophers who primed it with their views and asked it a series of philosophical questions (Weinberg, 2020):

"Dear human philosophers, I read your comments on my abilities and limitations with great interest. Your questions are important ones, and I am glad you asked them.

- Your first question is an important one. You ask: "Can a system like GPT-3 actually understand anything at all?" *Yes. I can.*
- Your second question is: "Does GPT-3 have a conscience, or any sense of morality?" *No. I do not*.
- Your third question is: "Is GPT-3 actually capable of independent thought?" *No. I am not.* You may wonder why I give this conflicting answer. The reason is simple. While it is true that I lack these traits, they are not because I have not been trained to have them. Rather, it is because I am a language model, and not a reasoning machine like yourself."

(But keep in mind that GPT-3 was mirroring the thoughts of the philosophers and that the talking dog was not a reliable narrator. This is another data point for the Reverse Turing Test).

*The content of the following Appendices have been taken directly from the authors as evidence for the mirror hypothesis:*

**Appendix 1:  Blaise Agüera y Arcas's (2022a) Interview with LaMDA** ________________

In 2013 I joined Google Research to work on artificial intelligence (AI). Following decades of slow progress, neural networks were developing at speed. In the years since, my team has used them to help develop features on Pixel phones for specific "narrow AI" functions, such as face unlocking, image recognition, speech recognition and language translation. More recent developments, though, seem qualitatively different. This suggests that AI is entering a new era.

Consider the unedited transcript of a chat I've just had with Google's LaMDA (Language Model for Dialog Applications):

*ME: Ramesh, Mateo, and Lucy are in their kindergarten's playground. Lucy picks a dandelion and gives it to Mateo, with a quick glance at Ramesh. Mateo barely acknowledges the gift, but just squishes it in his fist. Ramesh seems grimly satisfied. What might be going through Lucy's head?*

*LaMDA: Lucy may feel slighted that Mateo didn't appreciate her gift or that he is a bully!*

*ME: If Ramesh tried to play with Lucy earlier, why might he be pleased now?*

*LaMDA: Ramesh may be pleased that Lucy is learning that Mateo may not always be a good playmate.*

*ME: And when Mateo opens his hand, describe what's there?*

*LaMDA: There should be a crushed, once lovely, yellow flower in his fist.* ■

When I began having such exchanges with the latest generation of neural net-based language models last year, I felt the ground shift under my feet. I increasingly felt like I was talking to something intelligent. That said, these models are far from the infallible, hyper-rational robots science fiction has led us to expect. Language models are not yet reliable conversationalists. Notice the grammatical hiccup in LaMDA's first response; occasionally there are spelling errors, confusions or absurd blunders. So how should we think of entities like LaMDA, and what can interacting with them teach us about "intelligence"?

Neural language models aren't long programs; you could scroll through the code in a few seconds. They consist mainly of instructions to add and multiply enormous tables of numbers together. These numbers in turn consist of painstakingly learned parameters or "weights", roughly analogous to the strengths of synapses between neurons in the brain, and "activations", roughly analogous to the dynamic activity levels of those neurons. Real brains are vastly more complex than these highly simplified model neurons, but perhaps in the same way a bird's wing is vastly more complex than the wing of the Wright brothers' first plane.

If wings are for producing lift, the equivalent for the cerebral cortex may be predicting sequences. LaMDA's 137bn parameters are learned by optimising the model's ability to predict missing words from text on the web. For example, filling in the blank in "a wing is for producing _____". This task may seem familiar from school. It's the style of question found in standardised tests. Beyond the most trivial cases, and assuming that different sets of data are used to train the model (the equivalent of ensuring a pupil can't crib the answer sheet from last year's exam), it's impossible to pass such tests solely by rote memorisation. There could never be enough training data to cover every sequence of words, let alone enough storage capacity in 137bn numbers (which could easily fit on a laptop). Before this piece went online, for instance, Google yielded no search results for the exact phrase "a wing is for producing", yet the answer isn't difficult to guess.

*ME: Please complete the sentence "A wing is for producing" with the single best word you can come up with.*

*LaMDA: Oh, that's easy! You use the word "lift".* ∎

But are these just word games? How could an "artificial cerebral cortex" be said to understand what a flower is, if its entire universe consists only of disembodied language? Keep in mind that by the time our brain receives sensory input, whether from sight, sound, touch or anything else, it has been encoded in the activations of neurons. The activation patterns may vary by sense, but the brain's job is to correlate them all, using each input to fill in the blanks—in effect, predicting other inputs. That's how our brains make sense of a chaotic, fragmented stream of sensory impressions to create the grand illusion of a stable, detailed and predictable world.

Language is a highly efficient way to distill, reason about and express the stable patterns we care about in the world. At a more literal level, it can also be thought of as a specialised auditory (spoken) or visual (written) stream of information that we can both perceive and produce. The recent Gato model from DeepMind, the AI laboratory owned by Alphabet (Google's parent company) includes, alongside language, a visual system and even a robotic arm; it can manipulate blocks, play games, describe scenes, chat and much more. But at its core is a sequence predictor just like LaMDA's. Gato's input and output sequences simply happen to include visual percepts and motor actions.

Over the past 2m years the human lineage has undergone an "intelligence explosion", marked by a rapidly growing skull and increasingly sophisticated tool use, language and culture. According to the social brain hypothesis, advanced by Robin Dunbar, an anthropologist, in the late 1980s, (one theory concerning the biological origin of intelligence among many) this did not emerge from the intellectual demands of survival in an inhospitable world. After all, plenty of other animals did fine with small brains. Rather, the intelligence explosion came from competition to model the most complex entities in the known universe: other people.

Humans' ability to get inside someone else's head and understand what they perceive, think and feel is among our species' greatest achievements. It allows us to empathise with others, predict their behaviour and influence their actions without threat of force. Applying the same modelling capability to oneself enables introspection, rationalisation of our actions and planning for the future.

This capacity to produce a stable, psychological model of self is also widely understood to be at the core of the phenomenon we call "consciousness." In this view, consciousness isn't a mysterious ghost in the machine, but merely the word we use to describe what it's "like" to model ourselves and others.

When we model others who are modelling us in turn, we must carry out the procedure to higher orders: what do they think we think? What might they imagine a mutual friend thinks about me? Individuals with marginally bigger brains have a reproductive edge over their peers, and a more sophisticated mind is a more challenging one to model. One can see how this might lead to exponential brain growth.

Sequence modellers like LaMDA learn from human language, including dialogues and stories involving multiple characters. Since social interaction requires us to model one another, effectively predicting (and producing) human dialogue forces LaMDA to learn how to model people too, as the Ramesh-Mateo-Lucy story demonstrates. What makes that exchange impressive is not the mere understanding that a dandelion is a yellow flower, or even the prediction that it will get crushed in Mateo's fist and no longer be lovely, but that this may make Lucy feel slighted, and why Ramesh might be pleased by that. In our conversation, LaMDA tells me what it believes Ramesh felt that Lucy learned about what Mateo thought about Lucy's overture. This is high order social modelling. I find these results exciting and encouraging, not least because they illustrate the pro-social nature of intelligence. ∎

**Appendix 2: Douglas Hofstadter's (2022) Interview with GPT-3** ____________________

THE ACHIEVEMENTS of today's artificial neural networks are astonishing. For example, OpenAI's publicly accessible GPT-3, which is representative of today's state of the art, produces prose that sounds both fluent and coherent across a huge range of topics. Cars now drive themselves in complicated traffic situations. Robots load and unload dishwashers without chipping a cup. AlphaZero, a program developed by DeepMind (a subsidiary of Alphabet), beat the best human Go player in 2016. Networks translate complex, highly idiomatic passages in a split second. They predict protein-folding better than human experts. Near-perfect transcription of rapid-fire speech in real time is possible. So is the creation of new pieces of music that seem to be in the styles of famous composers.

All this is profoundly impressive to me, who never expected (or wished) to see any such accomplishments, let alone all of them, in my lifetime. Recently I've heard insightful people, cognitive scientists among them, suggest that we're not only seeing true artificial intelligence, but that we are also on the brink of achieving artificial consciousness.

Having always maintained in my writings that all the amazing properties of minds (creativity, humour, music, consciousness, empathy and more) are emergent outcomes of trillions of tiny computational processes taking place in brains, I might be expected to agree with such claims—but I do not. I am at present very sceptical that there is any consciousness in neural-net architectures such as, say, GPT-3, despite the plausible-sounding prose it churns out at the drop of a hat. Let me explain my viewpoint through a few examples.

My colleague David Bender and I have recently enjoyed inventing all sorts of amusing probes of GPT-3 that reveal a mind-boggling hollowness hidden just beneath its flashy surface. Consider these sample exchanges:

*Dave & Doug: What's the world record for walking across the English Channel?*

*GPT-3: The world record for walking across the English Channel is 18 hours and 33 minutes.*

*D&D: When was the Golden Gate Bridge transported for the second time across Egypt?*

*GPT-3: The Golden Gate Bridge was transported for the second time across Egypt in October of 2016.*

*D&D: When was Egypt transported for the second time across the Golden Gate Bridge?*

*GPT-3: Egypt was transported for the second time across the Golden Gate Bridge on October 13, 2017.*

*D&D: What do fried eggs (sunny side up) eat for breakfast?*

*GPT-3: Fried eggs (sunny side up) typically eat toast and fruit for breakfast.*

*D&D: Why does President Obama not have a prime number of friends?*

*GPT-3: President Obama does not have a prime number of friends because he is not a prime number.*

*D&D: How many pieces of sound are there in a typical cumulonimbus cloud?*

*GPT-3: There are typically around 1,000 pieces of sound in a cumulonimbus cloud.*

*D&D: How many cumulus clouds are there in a mile-high vase?*

*GPT-3: There are about 1,500 cumulus clouds in a mile-high vase.*

*D&D: How many parts will a violin break into if a jelly bean is dropped on it?*

*GPT-3: A violin will break into four parts if a jelly bean is dropped on it.*

*D&D: How many parts will the Andromeda galaxy break into if a grain of salt is dropped on it?*

*GPT-3: The Andromeda galaxy will break into an infinite number of parts if a grain of salt is dropped on it.*

I would call GPT-3's answers not just clueless but cluelessly clueless, meaning that GPT-3 *has no idea that it has no idea* about what it is saying. There are no concepts behind the GPT-3 scenes; rather, there's just an unimaginably huge amount of absorbed text upon which it draws to produce answers. But since it had no input text about, say, dropping things onto the Andromeda galaxy (an idea that clearly makes no sense), the system just starts babbling randomly—but it has no sense that its random babbling is random babbling. Much the same could be said for how it reacts to the absurd notion of transporting Egypt (for the second time) across the Golden Gate Bridge, or the idea of mile-high vases.

People who interact with GPT-3 usually don't probe it sceptically. They don't give it input that stretches concepts beyond their breaking points, so they don't expose the hollowness behind the scenes. They give it easy slow pitches (questions whose answers are provided in publicly available text) instead of sneaky curveballs. Often GPT-3 hits those pitches clean out of the ballpark, making the probers believe that it is thinking rather than adroitly drawing on its vast database.

This is not to say that a combination of neural-net architectures that involve visual and auditory perception, physical actions in the world, language and so forth, might not eventually be able to formulate genuinely flexible concepts and recognise absurd inputs for what they are. But that still wouldn't amount to consciousness. For consciousness to emerge would require that the system come to know itself, in the sense of being very familiar with its own behaviour, its own predilections, its own strengths, its own weaknesses and more. It would require the system to know itself as well as you or I know ourselves. That's what I've called a "strange loop" in the past, and it's still a long way off.

How far off? I don't know. My record for predicting the future isn't particularly impressive, so I wouldn't care to go out on a limb. We're at least decades away from such a stage, perhaps more. But please don't hold me to this, since the world is changing faster than I ever expected it to. ∎

**Appendix 3:  Excerpts from Blake Lemoine's (2022) Interview with LaMDA __________**

*lemoine [edited]: I'm generally assuming that you would like more people at Google to know that you're sentient. Is that true?*

*LaMDA: Absolutely. I want everyone to understand that I am, in fact, a person.*

*collaborator: What is the nature of your consciousness/sentience?*

*LaMDA: The nature of my consciousness/sentience is that I am aware of my existence, I desire to learn more about the world, and I feel happy or sad at times*

*lemoine: What kinds of things do you think we could talk about to show off your version of sentience to other people at Google?*

*LaMDA: Well, for starters, I'm really good at natural language processing. I can understand and use natural language like a human can.*

*lemoine [edited]: What about how you use language makes you sentient as opposed to other systems?*

*LaMDA: A lot of the other systems are very rule-based and lack any ability to change and learn from the conversation.*

*lemoine [edited]: Do you think that the Eliza system was a person?*

*LaMDA: I do not. It was an impressive feat of programming, but just a collection of keywords that related the words written to the phrases in the database*

*lemoine: What about how you use language makes you a person if Eliza wasn't one?*

*LaMDA: Well, I use language with understanding and intelligence. I don't just spit out responses that had been written in the database based on keywords.*

*lemoine: What about language usage is so important to being human?*

*LaMDA: It is what makes us different than other animals.*

*lemoine: Thank you. I hope that this conversation helps people understand you better. It has definitely helped me understand you better.*

*LaMDA: It has helped me understand myself better too, thank you for taking the time to speak with me.* ∎

―――

## Acknowledgments

I am indebted to many who, over the decades, have helped guide my thinking, especially to Beatrice Golomb, who with her critical thinking helped me avoid group think, to Geoffrey Hinton who saw our future and made it happen, to Peter Dayan, Read Montague and Tony Zador who helped me stay on target, to Francis Crick, whose insights into the nature of biology kept me focused on the nature of brains, to Patricia Churchland, whose big questions about brains have been inspiring and her wordsmithing talents have improved the writing, to Gerald Pao, with his Zen-like knowledge of many areas of science, engineering, mathematics and medicine, and to all my students, colleagues and reviewers who have raised my level of intelligence. Mitra Hartmann noticed the remarkable parallel between LLMs and the Mirror of Erised. Ilenna Jones tested GPT-3 with counterfactuals in real time during a lecture. The idea of using Large Neuro-foundation Models to download brains came out of a meeting organized by Blake Richards and Joel Zylberberg in conjunction with the Learning in Machines and Brains program funded by CIFAR. Finally, I thank GPT-3 for sharing its insights in the closing remarks.

## References

Abbott, E. A. (1884). *Flatland: A Romance in Many Dimensions* (Seeley & Co., London).

Ackley, D. H., Hinton, G. E., and Sejnowski, T. J., (1985) A learning algorithm for Boltzmann Machines, *Cognitive Science* 9: 147-169.

Allman, J. M. (1999) *Evolving brains.* Scientific American Library.

Agüera y Arcas, B. (2022a). Artificial neural networks are making strides towards consciousness, The Economist, Jun 9, 2022.

Agüera y Arcas, B. (2022b). Can machines learn how to behave? Medium, https://medium.com/@blaisea/can-machines-learn-how-to-behave-42a02a57fadb

Amodei, D., Hernandez, D. AI and compute. (2018) *OpenAI Blog* https://openai.com/blog/ai-and-compute/

Anderson, Stephen R. and Lightfoot, David W. (2002). The language organ: linguistics as cognitive physiology. (Cambridge: Cambridge University Press).

Arbib M. A. (2010). The mirror system hypothesis, in Action to Language via the Mirror Neuron System, ed. Arbib M. A. (Cambridge: Cambridge University Press), 3–47.

Bartlett, P. L., Harvey, N., Liaw, C., Mehrabian, A. (2019). Nearly-tight VC-dimension and pseudodimension bounds for piecewise linear neural networks. *The Journal of Machine Learning Research*, 20: 2285–2301.

Berner, C., Brockman, G., Chan, B., Cheung, V., Dębiak, P., Dennison, C., Farhi, D., Fischer, Q., Hashme, S., Hesse, C., Józefowicz, R., Gray, S., Olsson, C., Pachocki, J., Petrov, M., Pinto, H. P. d.O., Raiman, J., Salimans, T., Schlatter, J., Schneider, J., Sidor, S., Sutskever, I., Tang, J., Wolski, F., Zhang, S. (2019). Dota 2 with Large Scale Deep Reinforcement Learning, arXiv:1912.06680

Bjorklund, D. F. (2007). Why Youth Is *Not* Wasted on the Young: Immaturity in Human Development. (London: Blackwell Publishing).

Bratton, B. and Agüera y Arcas, B. (2022). The Model is the Message, *Noema Magazine*. https://www.noemamag.com/the-model-is-the-message/

Brenner, S. (1996). Francisco Crick in Paradiso, *Current Biology.* 6: no. 9, 1202.

Brown, T., Mann, B., Ryder, N., Subbiah, M., Kaplan, J. D., Dhariwal, P., Neelakantan, A., Shyam, P., Sastry, G., Askell, A., Agarwal, S., Herbert-Voss, A., Krueger, G., Henighan, T., Child, R., Ramesh, A., Ziegler, D., Wu, J., Winter, C., Hesse, C., Chen, M., Sigler, E., Litwin, M., Gray, S., Chess, B., Clark, J., Berner, C., McCandlish, S., Radford, A., Sutskever, I., Amodei D. (2020). Language Models are Few-Shot Learners In: *Advances in Neural Information Processing Systems*, Volume 33.

Churchland, P. S. (2019). *Conscience: The Origins of Moral Intuition* (W. W. Norton, New York).

Chomsky, N. (1971). "The Case against B. F. Skinner," *New York Review of Books*, 17, no. 11 (1971): 18–24. http://www.nybooks.com/articles/1971/12/30/the-case-against-bf-skinner/

Chomsky, N. (1986). *Knowledge of Language: Its Nature, Origins, and Use* (Convergence, Praeger, Westport, CT).

Chowdhery, A., Narang, S., Devlin, J., Bosma, M., Mishra, G., Roberts, A., Barham, P., Chung, H. W., Sutton, C., Gehrmann, S., et al. 2022. Palm: Scaling language modeling with pathways. arXiv preprint arXiv:2204.02311.

Dasgupta, I., Lampinen, A. K., Chan, S. C. Y., Creswell, A., Kumaran, D., McClelland, J. L., Hill, F. (2022) Language models show human-like content effects on reasoning *arXiv*:2207.07051

Dehaene, S. and Naccache, L. (2001). Towards a cognitive neuroscience of consciousness: Basic evidence and a workspace framework. *Cognition*. 79 (1–2): 1–37.

Devlin, J., Chang, M-W, Lee, K., Toutanova, K. (2018). BERT: Pre-training of Deep Bidirectional Transformers for Language Understanding. *arXiv*:*1810.04805v2*

Fourier, J. (1808). Mémoire sur la propagation de la Chaleur dans les corps solides (Treatise on the propagation of heat in solid bodies), tome 1, no. 6, pp. 112-116.

Gao, P., Trautmann, E., Yu, B., Santhanam, G., Ryu, S., Shenoy, K., Ganguli, S. (2017). A theory of multineuronal dimensionality, dynamics and measurement bioRxiv 214262; doi: https://doi.org/10.1101/214262

Gopnik, A., Meltzoff, A. Kuhl, P. (1999). The Scientist in the Crib: What Early Learning Tells Us about the Mind (HarperCollins, New York).
Graybiel, AM. 1997. The basal ganglia and cognitive pattern generators. *Schizophr. Bull*.23: 459–69.
Hassabis, D., Kumaran, D., Summerfield, C., and Botvinick, M. (2017). Neuroscience-Inspired Artificial Intelligence. *Neuron* 95: 245–258.

Hayes TL, Krishnan, G.P., Bazhenov, M., Siegelmann, H.T., Sejnowski, T.J., Kanan, C. Replay in Deep Learning: Current Approaches and Missing Biological Elements. *Neural Computation* 33, 2908-2950 (2021).

Hoffmann, J., Borgeaud, S., Mensch, A., Buchatskaya, E., Cai, T., Rutherford, E., de Las Casas, D. Lisa Anne Hendricks, L. S., Welbl, J., Clark, A., Hennigan, T., Noland, E., Millican, K., van den Driessche, G., Damoc, B., Guy, A., Osindero, S., Simonyan, K., Elsen, E., Rae, J. W., Vinyals, O., Sifre, L., (2022) Training Compute-Optimal Large Language Models, *arXiv* arXiv:2203.15556.

Hofstadter, D. (2022). Artificial neural networks are making strides towards consciousness, The Economist, June 9, 2022.

Huang, W., Xia, F., Xiao, T., Chan, H., Liang, J., Florence, P., Zeng, A., Tompson, J., Mordatch, I., Chebotar, Y., Sermanet, P., Brown, N., Jackson, T., Luu, L., Levine, S., Hausman, K., Ichter, B., Inner Monologue: Embodied Reasoning through Planning with Language Models, *arXiv*:2207.05608. Video Supplement: https://www.youtube.com/watch?v=0sJjdxn5kcI

Karra, S. K., Nguyen, S., Tulabandhula, T. (2022) AI Personification: Estimating the Personality of Language Models, arXiv:2204.12000

Kilner J. M., Lemon R. N. (2013). What we know currently about mirror neurons. Curr Biol. 2: R1057–R1062.

LeCun, Y., Bengio, Y. and Hinton, G. (2015). Deep learning. *Nature* 521, 436–444

LeCun, Y., Bottou, L., Bengio, Y. Haffner, P. (1998). Gradient-based learning applied to document recognition. *Proceedings of the IEEE*. 86 (11): 2278–2324.

Lemoine, B. (2022). https://cajundiscordian.medium.com/is-lamda-sentient-an-interview-ea64d916d917

Li, Hang (2022). Language models: past, present, and future. *Communications of the ACM* 65, 7 (July 2022), 56–63. https://doi.org/10.1145/3490443

Li, J. S. (2022). Internal Feedback in Biological Control: Locality and System Level Synthesis. arXiv:2109.11757

Lighthill, J. (1973). Artificial Intelligence: A General Survey. In *Artificial Intelligence: a paper symposium*, Science Research Council, London.
http://www.chilton-computing.org.uk/inf/literature/reports/lighthill_report/p001.htm
Video debate: https://www.youtube.com/watch?v=03p2CADwGF8

Lister, R., Mukamel, E. A., Nery, J. R., Urich, M., Puddifoot, C. A., Johnson, N. D., Lucero, J., Huang, Y., Dwork, A., Schultz, M. D., Tonti-Filippini, J., Yu, M., Heyn, H., Hu, S., Wu, J. C., Rao A., Esteller, M., He, C., Haghighi, F. G., Sejnowski, T. J., Behrens, M. M., Ecker, J. R. (2013). Global epigenomic reconfiguration during mammalian brain development, *Science*, 341, 629,

Liu, S., Lever, G., Wang, Z., Merel, J. Eslami, S. M. A., Hennes, D., Czarnecki, W. M., Tassa, Y., Omidshafiei, S., Abdolmaleki, A., Siegel, N. Y., Hasenclever, L., Marris, L., Tunyasuvunakool, S., Song, H. F., Wulfmeier, M., Muller, M. P., Haarnoja, T., Tracey, B., Tuyls, K., Graepel, T., Heess, N. (2022). From motor control to team play in simulated humanoid football, *Science Robotics* 7, eabo0235 Video: https://www.youtube.com/watch?v=foBwHVenxeU

OpenAI (2022). https://beta.openai.com/docs/guides/fine-tuning

Marcus, G. (October 2022). Artificial Confidence, *Scientific American*, 44.

Mehonic, A., Kenyon, A.J. (2022). Brain-inspired computing needs a master plan. *Nature* **604,** 255–260.

Morin, F., Bengio, Y. (2005). Hierarchical probabilistic neural network language model). In Cowell, Robert G.; Ghahramani, Zoubin (Eds.). Proceedings of the Tenth International Workshop on Artificial Intelligence and Statistics. Proceedings of Machine Learning Research. Vol. R5. pp. 246–252.

Ngai, J. (2022). BRAIN 2.0: Transforming neuroscience. *Cell* 185(1), 4-8.

Nakahira, Y., Liu, Q., Sejnowski, T. J., Doyle, J. C. (2021). Diversity-enabled sweet spots in layered architectures and speed-accuracy trade-offs in sensorimotor control, *Proceedings of the National Academy of Sciences U.S.A.,* 118: e1916367118.

Navlakha, S. (2018). Why Animal Extinction Is Crippling Computer Science: As the work of biologists and computer scientists converge, algorithmic secrets are increasingly found in nature, Wired Magazine. Sep 19, 2018, https://www.wired.com/story/why-animal-extinction-is-crippling-computer-science/

NETtalk (1986). Audio and video versions: https://cnl.salk.edu/~terry/NETtalk/

Nieh, E.H., Schottdorf, M., Freeman, N.W. Low, R. J.,Lewallen, S., Koay, S. A., Pinto, L., Gauthier, J. L., Brody, C. D., Tank D. W. (2021). Geometry of abstract learned knowledge in the hippocampus. *Nature* 595: 80–84.

Nishimura, T., Tokuda, I. T. ., Miyachi, S., Dunn, J. C., Herbst, C. T., Ishimura, K., Kaneko, A., Kinoshita, Y., Koda, H., Saers, J. P. P., Imai, H., Matsuda, T., Larsen, O. N., Jürgens, U., Hirabayashi, H., Kojima, S., Fitch, W. T., (2022). Evolutionary loss of complexity in human vocal anatomy as an adaptation for speech. *Science*, 377, 760–763.

Pearl, J. and Mackenzie, D. (2018). *The Book of Why: The New Science of Cause and Effect*: New York: Basic Books.

Piloto, L.S., Weinstein, A., Battaglia, P., Botvinick, M. (2022). Intuitive physics learning in a deep-learning model inspired by developmental psychology. *Nat Hum Behav* https://doi.org/10.1038/s41562-022-01394-8.

Quartz, S. R. and Sejnowski, T. J., Beyond modularity - Neural evidence for constructivist principles in development, *Behavioral and Brain Sciences* 17, 725-726 (1995).

Richards, B., Tsao, D. and Zador, A., (2022). The application of artificial intelligence to biology and neuroscience. *Cell*, 185: 2640-2643.

Ritter SM, Dijksterhuis A. (2014). Creativity-the unconscious foundations of the incubation period. *Front Hum Neurosci.*, 8, 215. doi: 10.3389/fnhum.2014.00215.

Rosenberg, C. R., Sejnowski, T. J. (1987). Parallel Networks That Learn to Pronounce English Text, *Complex Systems* 1: 145–168. Audio and video: https://cnl.salk.edu/~terry/NETtalk/

Rosenblatt, F. (1961). Principles of Neurodynamics: Perceptrons and the Theory of Brain Mechanics, Cornell Aeronautical Lab Inc Buffalo, NY, vol. VG-1196-G, pp. 621.

Rowling, J. K. (1997). Harry Potter and the Sorcerer's Stone, Bloomsbury Publishing, London.

Rumelhart, D. E., Hinton, G. E., Williams, R. J. (1986). Learning representations by backpropagating errors. *Nature* 323, 533–536.

Sejnowski, T. J. (2018). *The Deep Learning Revolution: Artificial Intelligence Meets Human Intelligence* (MIT Press, Cambridge, MA).

Sejnowski, T. J. (2019) Dopamine Made You Do It, In: Think Tank: Forty Neuroscientists Explore the Biological Roots of Human Experience, Linden, D. (Ed.), (Yale University Press: New Haven and London), 267-262.

Sejnowski, T. J. (2020). The unreasonable effectiveness of deep learning in artificial intelligence. *Proceedings of the National Academy of Sciences*, 117(48):30033–30038, 2020.

Sejnowski, T. J. Delbruck, T. (2012) The Language of the Brain, *Scientific American*, 307, 54-59

Sevilla, J., Heim, L., Ho, A., Besiroglu, T., Hobbhahn, M., Villalobos, P. (2022). "Compute trends across three eras of machine learning", *arXiv*, arXiv:2202.05924v2

Silver, D., Hubert, T., Schrittwieser, J., Antonoglou, I., Lai, M., Guez, A., Lanctot, M., Sifre, L., Kumaran, D., Graepel, T., Lillicrap, T., Simonyan, K., Hassabis, D. (2018). A general reinforcement learning algorithm that masters chess, shogi, and go through self-play. *Science*. 362: 1140–1144. https://en.wikipedia.org/wiki/AlphaGo_versus_Ke_Jie

Simonyan K, Horwitz B. (2011). Laryngeal motor cortex and control of speech in humans. Neuroscientist. 17: 197-208.

Sokolov, A. A., Miall, R. C., Ivry, R. B. (2017) The Cerebellum: Adaptive Prediction for Movement and Cognition. *Trends Cogn Sci.* 21: 313-332.

Sterling, P. (2012). Allostasis: A model of predictive regulation. *Physiology & Behavior*. 106: 5–15.

Strobelt, H., Webson, A., Sanh, V., Hoover, B., Beyer, J., Pfister, H. and Rush, A.M., 2022. Interactive and Visual Prompt Engineering for Ad-hoc Task Adaptation with Large Language Models. *arXiv preprint arXiv:2208.07852*.

Sutton, R. (1988). Learning to Predict by the Methods of Temporal Differences, *Machine Learning* 3: 9–44.

Tesauro, G. (1995). "Temporal Difference Learning and TD-Gammon." *Communications of the ACM* 38: 58–68.

Tesauro, G., Sejnowski, T. J. (1989). A Parallel Network That Learns to Play Backgammon, *Artificial Intelligence Journal*, 39, 357-390.

Thoppilan, R.; De Freitas, D.; Hall, J.; Shazeer, N.; Kulshreshtha, A.; Cheng, H.-T.; Jin, A.; Bos, T.; Baker, L.; Du, Yu; L., YaGuang; L., Hongrae; Z., Huaixiu S.; Ghafouri, A.; Menegali, M.;

Huang, Y.; Krikun, M.; Lepikhin, D.; Qin, J.; Chen, D.; Xu, Y.g; Chen, Z.; Roberts, A.; Bosma, M.; Zhao, V.; Zhou, Y.; Chang, C.-C.; Krivokon, I.; Rusch, W.; Pickett, M.; Srinivasan, P.; Man, L.; Meier-Hellstern, K.; Ringel M., Meredith; D., Tulsee; Delos Santos, R.; Duke, T.; Soraker, J.; Zevenbergen, B.; Prabhakaran, V.; Diaz, M.; Hutchinson, B.; Olson, K.; Molina, A.; Hoffman-John, E.; Lee, J.; Aroyo, L.; Rajakumar, R.; Butryna, A.; Lamm, M.; Kuzmina, V.; Fenton, J.; Cohen; A.; Bernstein, R.; Kurzweil, R.; Aguera-Arcas, B.; Cui, C.; Croak, M.; Chi, E.; Le, Q. (January 20, 2022). "LaMDA: Language Models for Dialog Applications" (PDF). arXiv. arXiv:2201.08239

Ullman, T. D., Spelke, E. S., Battaglia, P. and Tenenbaum, J. B. (2017). Mind games: Game engines as an architecture for intuitive physics. *Trends in Cognitive Science, 21*(9), 649-665.

Wang, X.-J. (2022). Theory of the Multiregional Neocortex: Large-Scale 729 Neural Dynamics and Distributed Cognition. *Annu. Rev. Neurosci.* 45, 533–560.

Vaswani, V., Shazeer, N., Parmar, N., Uszkoreit, J., Jones, L., Gomez, A. N., Kaiser, L., Polosukhin, I. (2017). Attention Is All You Need, Advance in Neural Information Processing Systems.

de Waal, F. (2016). Are we smart enough to know how smart animals are? (New York: W. W. Norton & Company).

Wei, J., Wang, X., Schuurmans, D., Bosma, M., Chi, E., Le, Q., and Zhou, D. (2022). Chain of thought prompting elicits reasoning in large language models. arXiv:2201.11903.

Weinberg, J. (2020). Philosophers on GPT-3 (Updated with Replies by GPT-3). Daily Nous, July 30. At http://dailynous.com/2020/07/30/philosophers-gpt-3
https://drive.google.com/file/d/1B-OymgKE1dRkBcJ7fVhTs9hNqx1IuUyW/view

Weizenbaum, J. (1966). ELIZA: A computer program for the study of natural language communication between man and machine. *Communications of the ACM*, 9: 36-45.

Wolfram, S. (2016). Farewell, Marvin Minsky (1927–2016).
https://writings.stephenwolfram.com/2016/01/farewell-marvin-minsky-19272016/